\documentclass[daat]{ipart}

\RequirePackage{hyperref}

\usepackage{tikz-cd}
\usetikzlibrary{decorations.shapes}

\usepackage{xcolor}
\newcommand{\CORRECTION}[1]{\textcolor{black}{#1}}

\startlocaldefs
\theoremstyle{plain}

\endlocaldefs
\startlocaldefs
\theoremstyle{plain}
\newtheorem{proposition}{Theorem}[section]

\endlocaldefs

\pubyear{2026}
\volume{1}
\issue{3}
\Doi{10.4310/DAT.260831104503}
\Pubonline{August 31, 2026}
\firstpage{193}
\lastpage{204}
\arxiv{2604.23804}

\begin{document}

\begin{frontmatter}
\title[Reparameterization through coverings and topological weight priors]{Reparameterization through coverings and topological weight priors}

\begin{aug}
    \author{\inits{M.}\fnms{Maxim} \snm{Beketov}\thanksref{t2}\ead[label=e1]{maxbeketov@outlook.com}}
    \address{National Research University Higher School of Economics\\ Moscow, 119048, Russia\\
             \printead{e1}}  
             \and 
    \author{\inits{P.}\fnms{Pavel} \snm{Snopov}\ead[label=e2]{paul.snopov@utrgv.edu}}
    \address{The University of Texas Rio Grande Valley\\ Texas, 78520, USA\\
             \printead{e2}}

    \thankstext{t2}{Corresponding author.}
\end{aug}
\received{\sday{23} \smonth{6} \syear{2026}}

\begin{abstract}
We generalise the reparameterization trick (RT) applied in variational autoencoders (VAEs) letting these have latent spaces of non-trivial topology – i.e. that of base manifolds covered with other ones, on which some technique for RT is available. That is possible since covering maps are measurable – moreover, \CORRECTION{this allows to establish an inequality on KL-divergence between pushforward (PF) densities on the base latent manifold, bounding it with KL-divergence between pullbacks on the cover, in some cases making the KL-term of VAE’s ELBO analytically tractable, despite the topological non-triviality of the supporting latent manifold.} Our development follows a route close but somewhat alternative to reparameterization on Lie groups, the latest proposal for which is to reparameterize PFs of normal densities from the Lie algebra – “through” the exponential map, seen by us as a particular case of what we propose to call reparameterization via covering (RVC). We demonstrate the working of our approach by constructing a VAE with the latent space of Klein bottle (not a Lie group) topology, which we call KleinVAE, successfully learning an appropriate artificial dataset. We discuss potential applicability of such topology-informed generative models as weight priors in Bayesian learning, particularly for convolutional vision models, where said manifold was peculiarly shown to have some relevance.
\end{abstract}



\end{frontmatter}

\section{Introduction}
Natural data often possesses certain regularities, forcing it to concentrate near certain (latent) manifolds in the feature (representation) space -- that is the statement of the manifold hypothesis. If some data is a priori known to possess such a manifold structure, this topological inductive bias should ideally be held in mind when modeling the data distribution -- which can be done with e.g. variational autoencoders (VAEs). Constructing a VAE involves introducing a family of conditional probability distributions of data latent features, called variational posteriors (VPs), the parameters of which can be inferred with low variance from input data, using the technique called the reparameterization trick (RT) \citep{VAE-Kingma-Welling,Rezende-reparameterization-trick-second-reference}. 

As first introduced, RT implies that the VP is some normal distribution (conditioned on input data) on Euclidean space of the latents, the topology of which is trivial (in homotopy theory sense: $\mathbb{R}^n$ is simply-connected, as its fundamental group is trivial), potentially unsuitable for modeling certain types of data -- if the latent dimension is constrained, \CORRECTION{not allowing for an embedding of the latent manifold into Euclidean space.} In this context, in recent years RT has been successfully generalized for latent spaces of various non-trivial topologies: e.g. ($n$-dim hyper-) spheres $S^n$ in \cite{davidson2018hyperspherical} and Lie groups in \cite{Reparam-Lie-groups} -- e.g. $n$-dimensional tori $T^n$ and, in fact, Euclidean spaces: with $\mathbb{R}^n$ forming both a Lie group and a Lie algebra for itself.

Inspired by these elegant developments, this paper proposes an alternative view (grounded in point-set topology and measure theory) on RT in non-trivial topologies. \CORRECTION{Our contribution is} that as a model latent space, one can, in principle, consider any topological manifold (base of the covering) that can be covered with some other (cover) manifold, on the latter of which some technique for RT is available. In present work we consider the case when the cover is the Euclidean space $\mathbb{R}^n$, on which as a family of ``source'' VPs one can consider the family of normal distributions $\mathcal{N}(\mu,\Sigma)$ -- particular probability measures (PMs) on $\mathbb{R}^n$. These can be pushed-forward (PF) onto the base by the covering map to make the family of ``target'', model VPs on that ``target'' base manifold. 

The key fact making this legit is that if both the base and the cover are measurable spaces, a covering between them is a \textbf{measurable mapping}: the PFs of source VPs from the cover constitute PMs on the base, so one can do probabilistic modeling on the ``target'' base manifold (with the cover being its source ``proxy''). E.g. Monte-Carlo estimation of expectations is done by just sampling from the source VP on the ``proxy'' cover, mapping the sample point onto the base with the covering map, and then averaging. This is required e.g. for computing the evidence lower bound (ELBO) of VAE models, since it has a form of expectation w.r.t. some probability measure on the latent space. 

The claimed contribution of present work is that reparameterization via coverings (RVC) is totally possible, due to said measurability of any covering map (it always being a local homeomorphism, so preserving the topology of measurable Borel sets), allowing to compute expectations in the above way: since the basic idea of RT is in the decomposition of the corresponding (parameterized) stochastic (generative) map into parts: 1) purely stochastic sampling, but of some standard (un-parameterized) RV; and then 2) re-parameterization of said standard RV's sample point with a purely deterministic reparameterization map. \CORRECTION{Moreover, said measurability of covering maps allows us to establish an inequality on the KL-divergence between VPs on the base in terms of KL between their pullbacks on the cover, which in some particular cases produces an analytically tractable lower bound, allowing to mitigate Monte-Carlo estimation of KL term of the evidence lower bound (ELBO).}

\CORRECTION{Thus, with present work we show that said scheme of reparameterization via covering (RVC) can be performed on various non-trivial topologies, previously impossible, as we demonstrate on an example of the latent manifold being the Klein bottle.}

\section{The Klein bottle \& its topology in image data}

In present work we show how reparameterization can be done on a wide class of manifolds, that need not necessarily be Lie groups. As an minimal non-trivial example, we consider the Klein bottle. The Klein bottle $\mathcal{K}$ is a compact surface (manifold of dimension 2), that is non-orientable -- and thus not a Lie group (since any Lie group is parallelizable). \CORRECTION{We thus see it as an example challenging to prior approaches, but tamable to our proposed technique. We are also motivated by the appearance of this manifold in natural data, as described in this section.} 

$\mathcal{K}$ can be realized as a unit square $[0,1]^2$ with points on the edges identified:
\begin{equation}
    (x,0) \equiv (x,1),~~(0,y) \equiv (1,1\!-\!y)~~~\forall x,y\in[0,1],
    \label{eq:Klein-bottle-gluing}
\end{equation}
\CORRECTION{(vertical edges are identified with a flip)}. So $\mathcal{K}\cong[0,1]^2/\!\equiv$, is a quotient of $[0,1]^2$ by the above (\ref{eq:Klein-bottle-gluing}) equivalence relation. Notably, such realization of $\mathcal{K}$ can be covered by the 2-dimensional torus $T^2$ realized as a rectangle $[0,2]\!\times\![0,1]$ with opposing points on the sides symmetrically identified (see (\ref{eq:Klein-bottle-torus-as-quotient}) below), the explicit covering map reads:
\begin{equation}
f_{2\to1}(x,y)=\begin{cases}
(x,\,y)\quad\textrm{if}\quad x<1,\\
(x\!-\!1,\,1\!-\!y)\quad \textrm{otherwise}.
\end{cases}
\label{eq:Klein-bottle-covering-map-by-torus}
\end{equation}
This map (\ref{eq:Klein-bottle-covering-map-by-torus}) is a 2-sheet covering $f_{2\to 1}:T^2\to\mathcal{K}$, as any point $y\in\mathcal{K}$ has two pre-images $x_{1,2}=f^{-1}(y)\in T^2$ in the two parts (where $x\leq1$ or otherwise) of $T^2$. See Fig. \ref{fig:torus-klein-cover}: \CORRECTION{on the first sheet ($x<1$) the coordinates are as is, while on the second sheet ($x>1$) the vertical coordinate is flipped.}

\begin{figure}
    \centering
    \begin{tikzpicture}[scale=1.5, every node/.style={scale=0.9}]
        \coordinate (A1) at (0,0);
        \coordinate (B1) at (2,0);
        \coordinate (C1) at (0,1);
        \coordinate (D1) at (2,1);
        \coordinate (F1) at (1,0);
        \coordinate (G1) at (1,1);

        \coordinate (A2) at (3,0);
        \coordinate (B2) at (4,0);
        \coordinate (C2) at (3,1);
        \coordinate (D2) at (4,1);

        \draw[thick, shorten >=2pt, shorten <=2pt, ->] (A1) -- (B1);
        \draw[thick, shorten >=2pt, shorten <=2pt, ->] (C1) -- (D1);
        \draw[thick, shorten >=2pt, shorten <=2pt, ->>, dashed] (G1) -- (F1);
        \draw[thick, shorten >=2pt, shorten <=2pt, ->>] (A1) -- (C1);
        \draw[thick, shorten >=2pt, shorten <=2pt, ->>] (B1) -- (D1);

        \draw[thick, shorten >=2pt, shorten <=2pt, ->] (A2) -- (B2);
        \draw[thick, shorten >=2pt, shorten <=2pt, ->>] (D2) -- (B2);
        \draw[thick, shorten >=2pt, shorten <=2pt, ->] (C2) -- (D2);
        \draw[thick, shorten >=2pt, shorten <=2pt, ->>] (A2) -- (C2);

        \draw[thick, ->] (2.1, 0.5) -- (2.9, 0.5) node[midway, above] {\footnotesize $f$};
        \draw[thick, ->] (2.1, 0.5) -- (2.9, 0.5) node[midway, below] {\footnotesize $x_i \mapsto y$};

        (A1) -- (B1) node[midway, below=8pt] {\small Torus};

        (A2) -- (B2) node[midway, below=8pt] {\small Klein bottle};

        \foreach \point in {A1,B1,C1,D1,A2,B2,C2,D2}
        \fill[black] (\point) circle (1pt);

        \coordinate (x1) at (3.3, 0.7);
        \node[below, black] at (x1) {$y$};
        \coordinate (y1) at (0.3, 0.7);
        \node[below, black] at (y1) {$x_1$};
        \coordinate (y2) at (1.3, 0.3);
        \node[below, black] at (y2) {$x_2$};
        \fill[black] (x1) circle (1pt);
        \fill[black] (y1) circle (1pt);
        \fill[black] (y2) circle (1pt);
    \end{tikzpicture}
    \caption{Two-sheet covering, see (\ref{eq:Klein-bottle-covering-map-by-torus}), $f_{2\to1}: T^2 \to \mathcal{K}$ of the Klein bottle $\mathcal{K}$ by a torus $T^2$. Each $y\in\mathcal{K}$ has 2 pre-images $x_1,x_2\in T^2$.}
    \label{fig:torus-klein-cover}
\end{figure}

Just as $\mathcal{K}$ can be two-sheet covered by $T^2$, the torus itself can be covered by $\mathbb{R}^2$, 2D Euclidean space -- its quotient $\mathbb{R}^2/\!\equiv$, by the following equivalence:
\begin{equation}
    (x_1,y_1)\equiv(x_2,y_2)~~\Leftrightarrow~~
    \begin{cases}
        x_1=x_2 \pmod 2\\
        y_1=y_2 \pmod 1.
    \end{cases}
    \label{eq:Klein-bottle-torus-as-quotient}
\end{equation}
is a rectangle $[0,2]\!\times\![0,1]$ with the above symmetric identification of edge points. This gives a map
\begin{equation}
    f_{\mathbb{R}^2/\mathbb{Z}^2}(x,y)=((x~~\textrm{mod}~~2),\,(y~~\textrm{mod}~~1)),
    \label{eq:Klein-bottle-torus-covering-map}
\end{equation}
an infinite-sheet (any point on $T^2$ has infinitely many preimages on $\mathbb{R}^2$) covering $f_{\mathbb{R}^2/\mathbb{Z}^2}:\mathbb{R}^2\to T^2.$

Combining maps (\ref{eq:Klein-bottle-covering-map-by-torus}) and (\ref{eq:Klein-bottle-torus-covering-map}) provides an infinite-sheet covering of $\mathcal{K}$ by $\mathbb{R}^2$:
\begin{equation}
    f_{\mathbb{R}^2\to\mathcal{K}}=f_{2\to 1}\circ f_{\mathbb{R}^2/\mathbb{Z}^2}.
    \label{eq:Klein-bottle-Klein-by-Euclid-total-covering-map}
\end{equation}
Note that, by construction, this covering is just, sheet-wise, an identical projection (albeit flipping the orientation on a half of all sheets) -- the base $\mathcal{K}$ is covered by (infinitely many) sheets that are unit squares in $\mathbb{R}^2$, so an open ball $B_r(x)$ (Borel subset) neighborhood (of sufficiently small radius $r$) of any point $x\in\mathbb{R}^2$ is mapped onto an open ball of the same radius in $\mathcal{K}$ realized as $[0,1]^2$ (modulo edgepoints equivalence). This makes the covering $f_{\mathbb{R}^2\to\mathcal{K}}$ a very ``nice'' mapping -- a measurable one -- in fact, any covering is such, being a local homeomorphism.

Another reason we're interested in this particular manifold, $\mathcal{K}$ -- is due to its \CORRECTION{appearance} in natural data -- which is, \CORRECTION{amazingly}, not limited to images: Klein bottle topology was also found in \cite{Cyclo-octane,Geometric-anomaly-detection} in the energy landscape of conformations of cyclooctane molecules. In the limits of present work we only consider the appearance of $\mathcal{K}$ in natural images.

To model Klein bottle topology found in natural images, \cite{Topo-Conv-Layers} propose to consider a certain subset of the Gabor filters (known in computer vision) -- ones given by the following function:
\begin{equation}
        F[\theta_1, \theta_2](x,y) = \sin(\theta_2)\,t_{\theta}(x,y) + \cos(\theta_2)\,Q\left(t_{\theta}(x,y)\right)
        \label{eq:Gabor-Klein-filter-function}
\end{equation}
on the domain of a square $\{(x,y)\}=[-1,1]^2$, parameterized by two angles, $\theta_1,\theta_2$, where \CORRECTION{$t_{\theta_1}(x,y)=\cos(\theta_1)\,x+\sin(\theta_1)\,y\,$} is projection onto a line specified by $\theta_1$, and $Q$ is a certain degree-2 Chebyshev polynomial: $Q(t)=2t^2-1$. This function $F$ is clearly periodic in its parameter space $\Theta=\{\theta_1,\theta_2\}$, with period $[0,2\pi]^2$ -- but, moreover, satisfies the following condition:
\begin{equation}
    F[\theta_1,\theta_2](x,y)=F[\theta_1\!+\!\pi,-\theta_2](x,y),
\end{equation}
making the fundamental domain in $\Theta$ not \CORRECTION{a torus of $[0,2\pi]^2$ with periodic boundary condition}, but rather its subset $[0,\pi]\!\times\![0,2\pi]$ with Klein bottle topology. If $F$ is discretized on a grid, one gets what we refer to as Gabor-Klein filters.


To prove that Gabor-Klein filters indeed have the topology of $\mathcal{K}$, one can compute their persistent homology (PH) \citep{Zomorodian-Carlsson,weinberger2011persistent}. We do exactly that: by sampling $(\theta_1,\theta_2)\sim U[[0,2\pi]^2]$ uniformly from the (double) fundamental region of $F$, we obtain a sample of 500 filters $3\!\times\!3$ filters and compute its PH with Ripser software library \cite{ripser}. These 500 filters are just real $3\!\times\!3$ matrices that we flatten turning them into a point cloud in $\mathbb{R}^{3\times3}=\mathbb{R}^9$. In this space, Euclidean distance is a natural metric, coinciding with Frobenius norm of difference of the un-flattened matrices -- PH is homology of Vietoris-Rips filtration of its sub-level sets.

\begin{figure}
\includegraphics[width=\linewidth]{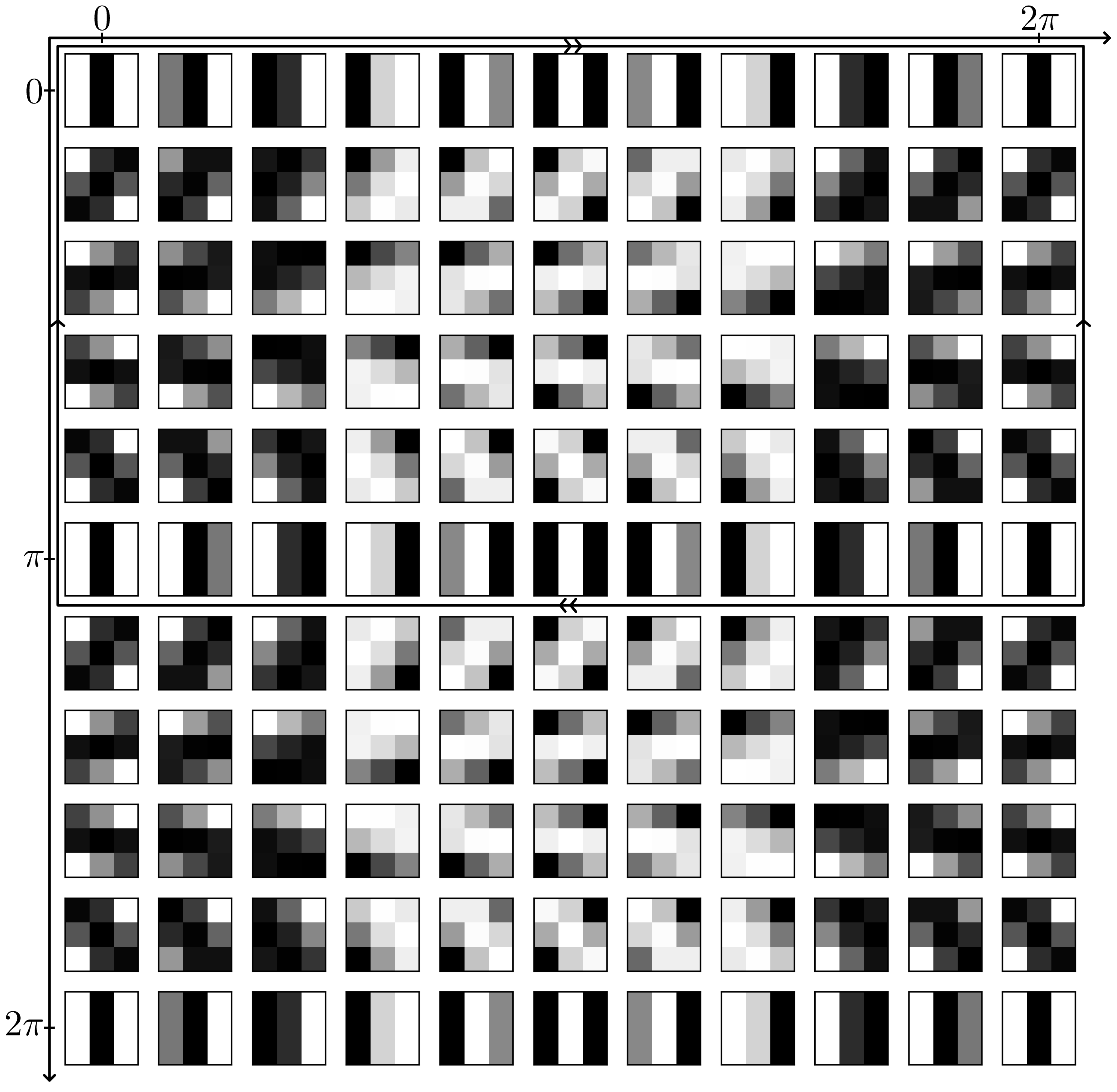} 
\caption{Gabor-Klein filters = discretizations of function $F$ (\ref{eq:Gabor-Klein-filter-function}), with $\theta_1,\!\theta_2$ params, on a $3\!\times\!3$ grid. Upper half, $\theta_1\!\leq\!\pi$, is the fund. region of $F$ in the parameter space $\Theta$, with Klein bottle topology. This is also an illustration of a 2-sheeted covering of $\mathcal{K}$ by a 2-torus: the whole square doubly covers the top half.}
\end{figure}

\begin{figure}
\includegraphics[width=\linewidth]{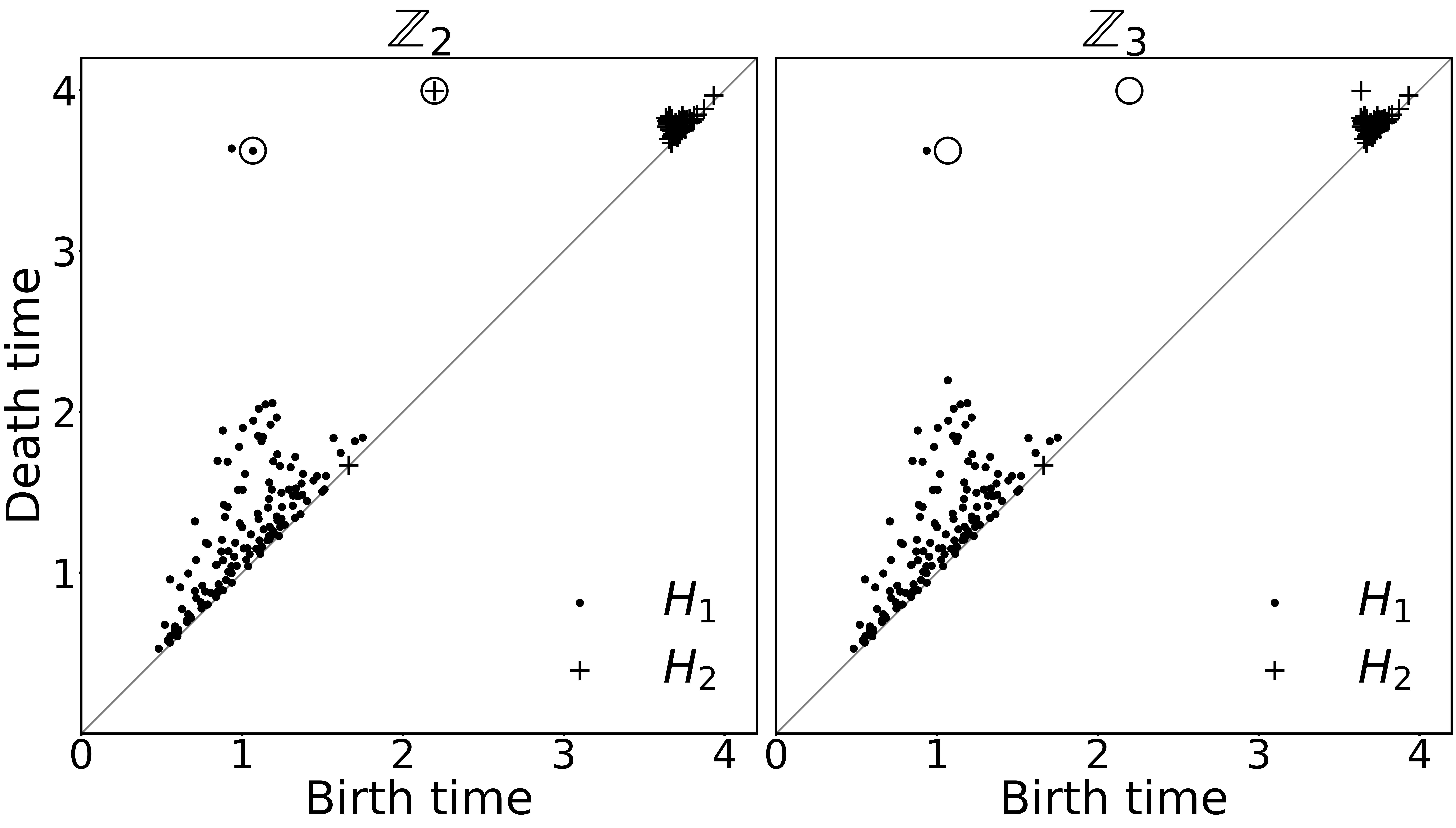}
\caption{Persistence diagrams of 500 filters sampled from $U[\Theta]$: left is PH over $\mathbb{Z}_2$; right is over $\mathbb{Z}_3$. Persistent features that exist over $\mathbb{Z}_2$ and disappear over $\mathbb{Z}_3$ are circled.}
\label{fig:diagrams-of-filters}
\end{figure}

Persistence diagrams we obtain are shown on Fig. \ref{fig:diagrams-of-filters}. We do not display 0-homology for visual clarity. Recall that reduced homology groups of the Klein bottle over $\mathbb{Z}$ are $H_0\cong 0,~H_1\cong\mathbb{Z}\oplus\mathbb{Z}_2$ and $H_2\cong0$ \citep{Hatcher}. So, if computed over $\mathbb{Z}_2$ and $\mathbb{Z}_3$, they become respectively:
\begin{equation}
\begin{array}{rl@{\quad}rl}
\begin{cases}
H_1(\mathcal{K};\mathbb{Z}_2) \cong \mathbb{Z}_2\oplus\mathbb{Z}_2, \\
H_2(\mathcal{K};\mathbb{Z}_2) \cong \mathbb{Z}_2,
\end{cases}
\begin{cases}
H_1(\mathcal{K};\mathbb{Z}_3)\cong\mathbb{Z}_3, \\
H_2(\mathcal{K};\mathbb{Z}_3)\cong0.
\end{cases}
\end{array}
\end{equation}

This explains Fig. \ref{fig:diagrams-of-filters}: over $\mathbb{Z}_2$, both components of $H_1$ are rightfully seen persisting, with one of them, originally $\mathbb{Z}$ over $\mathbb{Z}$, now reduced to $\mathbb{Z}_2$, yet still non-trivial. Over $\mathbb{Z}_3$, the other component of $H_1$,  $\mathbb{Z}_2$ from the start over $\mathbb{Z}$ -- disappears, hence not persisting. The ``phantom'' 2-homology of $\mathbb{Z}_2$ over $\mathbb{Z}_2$ rightfully disappears over $\mathbb{Z}_2$. Higher homology groups of $\mathcal{K}$ are trivial, since it's a 2-manifold. Aforementioned effects are due to torsion of $\mathcal{K}$ being a non-orientable manifold. This raises the so-called field choice problem in computational homology \citep{obayashi2019field}, that was in fact shown solvable in \cite{boissonnat2019computing} by effectively computing PH over many fields at once. These torsion effects make the topology of $\mathcal{K}$ (that can only be truly spotted by computing PH over $\mathbb{Z}_2$ and $\mathbb{Z}_3$ simultaneously) a notable test case for topological data analysis, as compared to other similarly low-dimensional manifolds.

\section{Reparameterization in different topologies}
Probabilistic generative modeling on manifolds is a broad and developing subject, reviewing it whole is beyond the scope of present paper. We only focus on variational autoencoder (VAE) models, with the latent space having some required non-trivial topology (as opposed to that of $\mathbb{R}^n$, Euclidean space). The reasons for such choice are as follows. 

First: VAEs do not require an explicit parameterization of the latent manifold $\mathcal{Z}$, as a sufficiently expressive decoder (paired with a proper encoder) can itself learn a parameterization map embedding (if possible) $\mathcal{Z}$ back into the data generation space $\mathcal{X}$. All one needs is to implicitly set latent topology -- that of $\mathcal{Z}$, that is: by introducing a reparameterizeable family of variational posterior (VP) probability measures (conditioned on data observed in $\mathcal{X}$) on $\mathcal{Z}$. Specifically, said VP family should permit: 1) sampling from the VP distribution on $\mathcal{Z}$ (to compute expectations of measurable functions -- random variables -- on $\mathcal{Z}$ with Monte-Carlo), 2) computing the KL-divergence between the VP and the prior distribution of latent parameters -- coordinates on $\mathcal{Z}$ manifold. Both these computations should allow passage of their backward gradients w.r.t. parameters of VAE's layers, and, at that, the parameters of the VP distribution (as output of the encoder given sample points in $\mathcal{X}$ as input). Selecting VPs to be push-forwards, by some measurable map (covering being such), onto $\mathcal{Z}$ -- allows these possibilities. To sample a point from a VP distribution being a PF by the covering map, one just samples from its pullback probability measure on the ``proxy'' cover of $\mathcal{Z}$, then projecting the sample point onto it, the base, with the covering map. By definition of pushforward, probability measures of all random events are ``re-calculated'' by the PF function correctly. As for reparameterizing this sampling map, the output of which is further fed to the decoder on a forward pass; and as for computing the KL-divergence between the VP and the prior defined on $\mathcal{Z}$ -- both as PFs of their ``proxies'' (pullbacks) on the cover -- see further. \CORRECTION{As we show}, an analytic inequality on KL-divergence arises between two ``target'' distributions on $\mathcal{Z}$, that are PFs by such covering -- bounding it by the KL between their ``proxy'' pullbacks -- from the right side, so minimizing KL between pullbacks delivers minimizing it between the pushforwards by such a covering map. 

Second reason for consideration of present work being limited to VAEs is that despite, undoubtably, more expressive (density-based) topology-aware generative models were proposed, like e.g. normalizing flows in \cite{rezende2015variational} (which were generalized to have latent spaces of toric and spherical topology in \cite{rezende2020normalizing}, and potentially more general topologies in \cite{brehmer2020flows}), that are capable of learning multimodal distributions, VAEs can be seen as a basic building blocks for such models to figure out first. After all, the (un-) normalization map of flows can be seen as a more intricate development of the reparameterization map.

We see it that, with the proposed technique of reparameterization via coverings (RVC), detailed further, one has all the ingredients for a topological VAE (provided its model latent space has such topology that it admits a covering with another ``proxy'' manifold, whereon some technique for RT is already available). Such TopoVAEs should not be confused with topological authoencoders: TopoAEs as proposed in \cite{moor2020topological} and RTD-AEs in \cite{trofimov2023learning} -- in these, the ambient latent space is Euclidean, into which the model learns to embed a point cloud of informative latent codes conditioned on observed input data -- in such a way that the persistent homology these latent representations is close to that of input ones. Said closeness is measured with some (differentiable, at least almost everywhere, thus learnable) distance between the persistence diagrams, or as a norm of the diagram of a special cross-filtration (representation topology divergence of RTD-AEs). In contrast to this, the class of VAE models we propose initially has proper implicitly set latent topology. Also, making ``vanilla'' AEs data-generative is possible \citep{bengio2013generalized}, but is, sadly, an arguably underdeveloped direction of research; while presently proposed TopoVAEs are, architecturally, just VAEs, by-design generative models that are well understood.

\subsection{Euclidean reparameterization, Spheres}

The key to constructing a VAE is reparameterization \citep{VAE-Kingma-Welling,Rezende-reparameterization-trick-second-reference}. Consider it for the case when the latent space of a VAE is just the 1D Euclidean space, the real line $\mathbb{R}$. The reparameterization (setting parameters $\mu$ and $\sigma$) transform on it reads:
\begin{equation}
    z=f_{\mu,\sigma}(\varepsilon)=\mu+\sigma\cdot\varepsilon,
    \label{eq:reparameterization-1D}
\end{equation}
given as input a sample point $\varepsilon\sim\mathcal{N}(0,1)$ of a standard normal random variable (RV), $f_{\mu,\sigma}$ is a (deterministic) map that outputs such a sample point of a normal RV $z\sim\mathcal{N}(\mu,\sigma)$ with parameters $\mu,\sigma$. To construct a Topo-VAE with a topologically non-trivial latent space $\mathcal{Z}$, one needs to generalize (\ref{eq:reparameterization-1D}) to work on this space.

Note that (\ref{eq:reparameterization-1D}) exploits the the affine structure of $\mathbb{R}$. The real variable $\varepsilon$ that is input to $f_{\mu,\sigma}$ can be seen as a 1D vector, lying in $\mathbb{R}$. With $\mathbb{R}$ forming a vector space (over itsel, $\mathbb{R}$, as a base field), vectors in it can be added and multiplied by a scalar from the base field. This, in fact, is also a manifestation of 1D Euclidean space $\mathbb{R}$ being a Lie group $G$, with (\ref{eq:reparameterization-1D}) giving the affine group action on this manifold: a combination of affine maps is again an affine map, and it depends on its parameters $\mu,\sigma$ smoothly. Notably, $G$ also serves as a Lie algebra $\mathfrak{g}$ for itself -- being a vector space as above explained. The (Lie) exponential map from $\mathfrak{g}$ to $G$ here is just the identity map.

The above described view is exactly what was proposed (generalized to other non-trivial examples) in \cite{Reparam-Lie-groups} as reparameterization on manifolds that are Lie groups. It should be noted, however, that Lie group structure is only one of many possible structures a manifold can be equipped with (if not to say it's arguably one of the ``nicest'', hence most restrictive, structures) and one can potentially exploit those other ones to design reparameterization tricks.

For example, if the latent manifold $\mathcal{Z}$ is Riemannian (equipped with a metric), one can do reparameterization by sampling random walks (diffusion) on it, as proposed in \cite{rey2019diffusion}: the intuition behind this is that in the limit, random walks converge to some form of Riemannian normal law with density at point $z$ proportional to $\exp(-d^2(\mu,z))$. There is a lot of progress in this direction of generative modeling on manifolds (see also \citep{gemici2016normalizing,yu2025learning,kalatzis2021density} for more flow-based approaches), but present work follows a different, more measure-theoretic, path (with a VAE as a basic vehicle): we do not explicitly require a Riemannian structure of the manifold.

Before considering Lie groups, let us briefly mention a very interesting case of (hyper-) spherical topologies -- separately, since not all spheres are Lie groups \citep{megia2007spheres}. Reparameterization can be generalized to spheres of arbitrary dimension \citep{davidson2018hyperspherical} exploiting other structures: e.g. using a probabilistic technique of reparameterization through accept-reject sampling on spheres was proposed in \cite{naesseth2017reparameterization}, only appealing to the compactness of spheres. We envision an alternative approach: computing implicit reparameterization gradients proposed in \cite{figurnov2018implicit} ``through'' smooth automorphisms of spheres. A very similar technique for spherical normalizing flows via Möbius transformations group was proposed in \cite{rezende2020normalizing}.

\subsection{Lie group density reparameterization}

Our work was inspired by reparameterization generalized to the case of the latent manifold being a Lie group \citep{falorsi2018explorations,Reparam-Lie-groups}.

\begin{figure}
\includegraphics[width=0.6\linewidth]{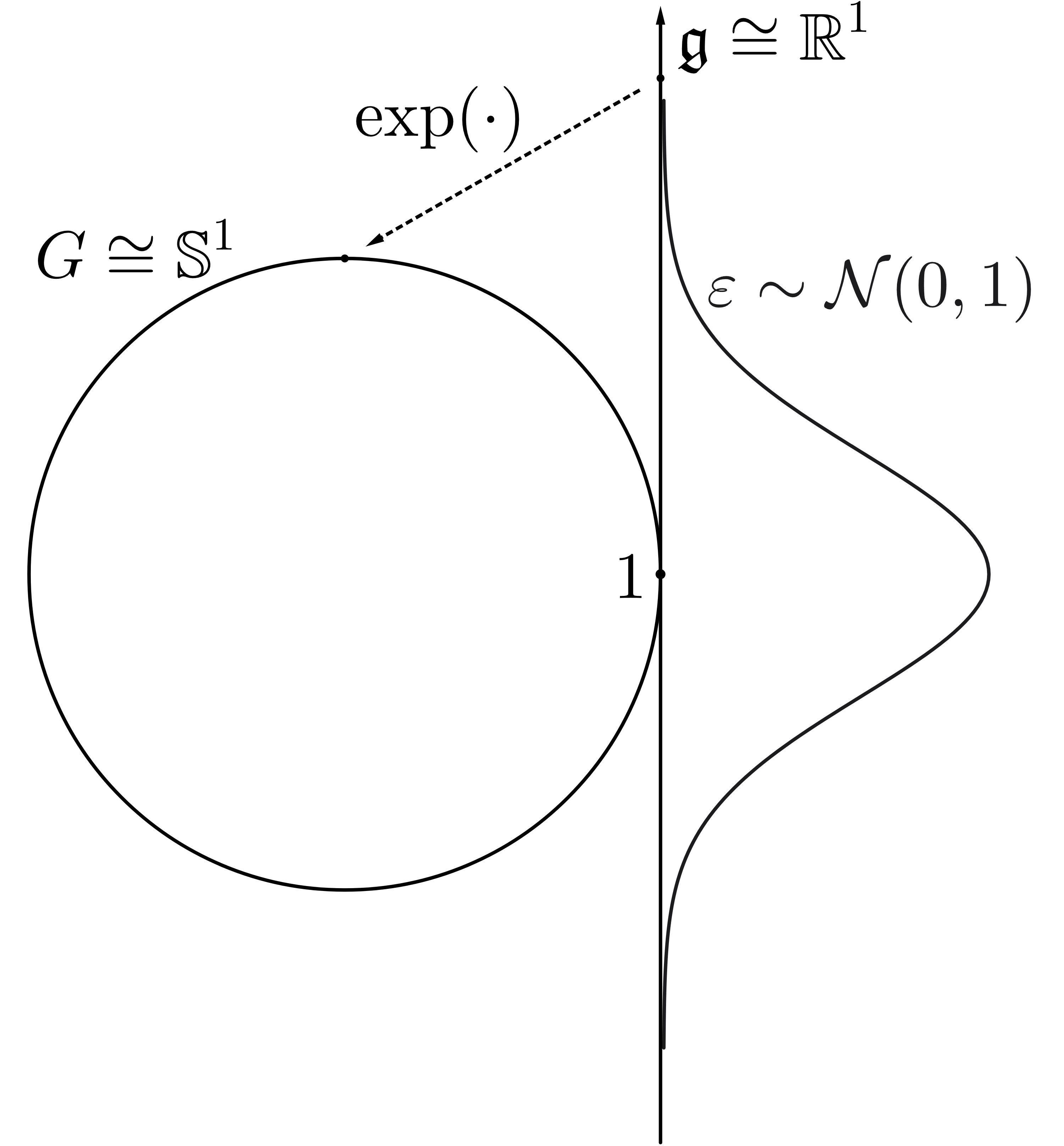} 
\caption{Lie-exp map from $\mathfrak{g}\cong \mathbb{R}^1$ to $G\cong \CORRECTION{\mathbb{S}^1}$, normal probability measure on $\mathfrak{g}$.}
\label{fig:circle}
\end{figure}

\CORRECTION{A Lie group $G$ is a smooth manifold that is also a group such that the multiplication and inversion operations are smooth}. Consider a 1-sphere $\CORRECTION{\mathbb{S}^1}$, embedded in Euclidean plane $\mathbb{R}^2$ with Cartesian coordinates, centered at its zero origin. Any point on $\CORRECTION{\mathbb{S}^1}$ can be parameterized with an angle $\varphi$ that one counts wrapping around the circle counterclockwise starting from point with coordinates $(1,0)$. A line tangent to the circle at this point $(1,0)$ (identity group action element of $G$) can be seen as a vector space $\mathbb{R}$ -- this is the Lie algebra $\mathfrak{g}$ of $G$. There is a smooth map from $\mathfrak{g}$ to $G$, called the exponential map, which in this example, given an angle $\varphi\in\mathfrak{g}$, returns a point on $\CORRECTION{\mathbb{S}^1}$ (in $G$) with coordinates $(\cos\varphi,\sin\varphi)$ in the plane, see Fig. \ref{fig:circle}.

To work with distributions on $\CORRECTION{\mathbb{S}^1}$, it being a Lie group $G$, one can consider distributions on $\mathfrak{g}\cong\mathbb{R}$ -- e.g. normal distributions. Any normal probability density $\mathcal{N}(\mu,\sigma)$ on $\mathfrak{g}\cong\mathbb{R}$ can be pushed onto $G\cong \CORRECTION{\mathbb{S}^1}$ by the exponential map -- its pushforward is called wrapped normal distribution (on the circle). The idea of reparameterization on Lie groups is that one can reparameterize distributions on Lie algebras of Lie groups (this works at least for compact connected ones \citep{falorsi2018explorations,Reparam-Lie-groups}) -- which are then pushed forward onto the group by the exponential map. So, the measurable space of the Lie algebra $\mathfrak{g}$ serves as a ``proxy'' of that of the Lie group $G$ it corresponds to: instead of working with probability densities on $G$, one can work with their $\mathfrak{g}$-supported ``proxies''. 

This reparameterization through the exp map onto Lie groups is, in fact, also a particular case of what we call reparameterization through the coveging map, since the above described exponential map from $\mathfrak{g}\cong\mathbb{R}$ to $G\cong \CORRECTION{\mathbb{S}^1}$ also delivers an infinite-sheet covering. Although not always, Lie-exp maps sometimes also serve as universal covering maps $\mathfrak{g}\to G$ (when $G$ is connected and abelian).

\CORRECTION{But covering maps are more general than Lie exp maps, and hence less restrictive. The latter are always smooth, while the former need not -- not expected to be more than local homeomorphisms, coverings may have arbitrarily complicated sets of discontinuities. As we show, when constructing a reparameterization map, this is not a problem if said set of discontinuities is of measure zero.}


\subsection{Covered spaces: circle, Klein bottle}

We outline a method to reparameterize probability measures on certain manifolds -- namely, ones that can be covered with other ones (on which one can do reparameterization). This is possible since any covering is a measurable map.  We do so on a particular example -- the Klein bottle, motivated by the mentioned appearance of this peculiar manifold in natural data.

\subsubsection{RVC on the circle of $
\CORRECTION{\mathbb{S}^1}$}
Topologically, $\CORRECTION{\mathbb{S}^1}$ can be realized as a quotient $\mathbb{R}/\mathbb{Z}$ of the real line $\mathbb{R}$ by the integer lattice $\mathbb{Z}$ -- by the following equivalence relation:
\begin{equation}
    x\equiv y\quad\Leftrightarrow \quad x=y\pmod 1,
    \label{eq:reparam-through-covering-circle-as-quotient}
\end{equation}
so the circle $\CORRECTION{\mathbb{S}^1}$ is isomorphic to an interval $[0,1]$ with endpoints identified, $0\!\equiv\!1$. The map
\begin{equation}
    f(x)= (x~~\textrm{mod}~~1)
    \label{eq:reparam-through-covering-circle-mod-map}
\end{equation}
(taking the fractional part of a number) thus provides a covering $f:\mathbb{R}\to[0,1]$ of the base interval with $\mathbb{R}$ (the cover). This is an infinite-sheet covering: every point $x\in[0,1]$ has (countably) infinitely many pre-images $f^{-1}(x)$ in the cover.

The real line $\mathbb{R}$ with a Borel sigma algebra $\mathcal{B}(\mathbb{R})$ on it, $(\mathbb{R},\mathcal{B}(\mathbb{R}))$ -- is a measurable space that can be equipped with a ``natural'' measure -- Lebesgue measure $\lambda$, with respect to which one can integrate measurable functions (compute expectations of RVs). Normal distributions $\mathcal{N}(\mu,\sigma)$ form a family of probability distributions (on $\mathbb{R}$) of very special interest -- these are often used as VPs. 

The distribution of a normal RV $X\!\sim\!\mathcal{N}(\mu,\sigma)$ is a probability measure $P_X$ on $\mathbb{R}$ that is absolutely continuous w.r.t. $\lambda$, so it has a probability density function (PDF) $p_X(x)$, the Gaussian:
\begin{equation}
    dP_X=p_X\,d\lambda=(2\pi\sigma^2)^{-1/2}e^{-(x-\mu)^2/2\sigma^2}\,d\lambda.
    \label{eq:reparam-through-covering-circle-normal-density}
\end{equation}

The covering (\ref{eq:reparam-through-covering-circle-mod-map}) $f:\mathbb{R}\to[0,1]$ is a measurable map from $(\mathbb{R},\mathcal{B}(\mathbb{R}),\lambda)$ to $[0,1]$ (with a Borel sigma-algebra on it), so one can consider the pushforward (PF) $f_*P_X$ of $P_X$ -- a measure on $[0,1]$. Pushforward measure of a small neighborhood $B=B_\varepsilon(y)$ of any point $y\in[0,1]$ is thus given by
\begin{equation}
    f_*P_X(B) = \sum_{\{A_i\}=f^{-1}(B)}P_X(A_i),
    \label{eq:reparam-through-covering-circle-pushforward-measure-of-ball-sum-over-preimages}
\end{equation}
where $A_i$ are small open neighborhoods of (infinitely many) points $\{x_i\}=f^{-1}(y)$ -- pre-images of $y$. In fact, sheet-wise (restricted to any sheet $\mathcal{X}_i\subset\mathbb{R}$, $f$ is a bijection), the covering map $f$ is just an identical projection, so:
\begin{equation}
    d(f_*P_X)(y)=\sum_{i} p_X(x_i)\,d\lambda=\big(\sum_ip_X(x_i)\big)\,d\lambda,
    \label{eq:reparam-through-covering-circle-dpushforward}
\end{equation}
it makes sense to talk about the pushforward density $f_*p_X$ (Radon-Nikodym derivative of $f_*P_X$ w.r.t. $\lambda$), which, at any point $y$, is given by the above \eqref{eq:reparam-through-covering-circle-dpushforward} infinite sum (that converges due to properties of the Gaussian) of normal densities $p_X$ at pre-images $\{x_i\}=f^{-1}(y)$. This distribution on the circle $\CORRECTION{\mathbb{S}^1}\cong([0,1]/\!\equiv)$ is referred to as wrapped normal distribution -- it inherits many properties of the normal on $\mathbb{R}$: in particular, it concentrates (the more the smaller the value of $\sigma$) around its single mode at $f(\mu)$, so it is naturally used to model circular RVs in directional statistics.

Now, consider two probability distributions on $\mathbb{R}$ -- with densities $q$ and $p$. If these are pushed onto $[0,1]$ by $f$ (denote pushforwards $q^*=f_*q$ and $p^*=f^*p$) and one considers Kullback–Leibler (KL-) divergence between the PFs:
\begin{equation}
    \textrm{KL}(q^*\,||\,p^*)=\int_{[0,1]}q^*\,\log\frac{q^*}{p^*}\,d\lambda
    \label{eq:reparam-through-covering-circle-KL}
\end{equation}
where both $q^*$ and $p^*$ at any point $y\in[0,1]$ equal to sums $q^*(y)=\sum_i q(x_i)=\sum_i q_i$, same for $p^*(y)=\sum_i p_i$ of pullback densities at pre-images $x_i$ of $y$. Then, by the Log Sum Inequality (following from Jensen's inequality):
\begin{equation}
\begin{aligned}
\int_{[0,1]}q^*\,\log\frac{q^*}{p^*}\,d\lambda=
\int_{[0,1]}\big(\sum_i\,q_i\big)\,\log\frac{\sum_i q_i}{\sum_i p_i}\,d\lambda\leq\\
\leq\int_{[0,1]}\sum_i q_i\,\log\frac{q_i}{p_i}\,d\lambda.
\label{eq:our-res-1}
\end{aligned}
\end{equation}
-- the integrand, and hence the integral, in (\ref{eq:reparam-through-covering-circle-KL}) is bounded from above by the KL-divergence between the original (pullback) densities $q$ and $p$ on $\mathbb{R}$. The above is because, if one permutes the sum and the integral in (\ref{eq:our-res-1}), one just gets a sum of integrals over all the sheets of the covering, which equals the integral over the whole $\mathbb{R}$:

\begin{proposition}
$\textrm{KL}_{\CORRECTION{\mathbb{S}^1}}(q^*\,||\,p^*)\leq \textrm{KL}_{\mathbb{R}^1}(q\,||\,p)$.
\end{proposition}

\subsubsection{RVC on the Klein bottle}


With that established, one can consider the case of Klein bottle $\mathcal{K}$ topology. As described in the introduction, $\mathcal{K}$ can be realized as a unit square $[0,1]^2$ with points on the sides appropriately identified. It can be 2-sheet covered (\ref{eq:Klein-bottle-covering-map-by-torus}) by a 2-torus, $T^2$, realized as a rectangle $[0,2]\times[0,1]$ with endpoints identified symmetrically as in (\ref{eq:Klein-bottle-torus-as-quotient}). Denote said covering map $f_{2\to1}$. Note that $f_{2\to1}$ is a sheet-wise identical projection map (albeit flipping the orientation on half of the sheets) -- so if $T^2\cong([0,2]\times[0,1]/\!\equiv)$ is equipped with Lebesgue measure $\lambda$, inherited from $\mathbb{R}^2$, then on each sheet $\lambda$ is preserved by $f_{2\to1}$. The torus, in turn, can be realized as a quotient of $\mathbb{R}^2$ by \CORRECTION{$2\mathbb{Z}\times\mathbb{Z}$} -- this provides a (infinite-sheet) covering, denoted $f_{\mathbb{R}^2/\mathbb{Z}^2}$, which is also, sheet-wise, just a truly identical projection preserving $\lambda$. So $\mathcal{K}\cong([0,1]^2/\!\equiv)$ is covered with $\mathbb{R}^2$ with a map 
\begin{equation}
f=f_{2\to1}\circ f_{\mathbb{R}^2/\mathbb{Z}^2}:\mathbb{R}^2\to\mathcal{K}. 
\label{eq:reparam-through-covering-Klein-covering-map}
\end{equation}
With that, one can introduce RVs on $\mathcal{K}$, by introducing RVs on $\mathbb{R}^2$ and then pushing their distributions onto $\mathcal{K}$ by $f$.

Similarly to the case of the circle, for $p^*$ and $q^*$ -- pushforward densities of $p$ and $q$ from $\mathbb{R}^2$ onto $\mathcal{K}$ by $f$, it holds true that:

\begin{proposition}
$\textrm{KL}_{\mathcal{K}}(q^*\,||\,p^*)\leq \textrm{KL}_{\mathbb{R}^2}(q\,||\,p)$.
\end{proposition}

In general, such an inequality should hold for any two manifolds
covering each other due to covering maps being measurable -- for measurable maps this is known in information theory as the data processing inequality (see e.g. Theorem 6.2 in \cite{polyanskiy2025information}).

Thus to construct a TopoVAE with its latent space topology that of $\mathcal{K}$, one can pick the PFs of normal distributions $\mathcal{N}(\mu,\Sigma)$ from $\mathbb{R}^2$ onto $\mathcal{K}$ by $f$ as a family of variational posteriors (VPs). PFs of $\mathcal{N}(\mu,\Sigma)$ onto $T^2$ by $f_{\mathbb{R}^2/\mathbb{Z}^2}$ are known in (directional) statistics as multivariate wrapped normal distributions. The unimodality of $\mathcal{N}(\mu,\Sigma)$-s may break when these are pushed onto $\mathcal{K}$ by $f$, but at least in the limit of small variance, when $\mathcal{N}(\mu,\Sigma)$ is close to a delta-measure at $\mu$, its PF by $f$ will also concentrate around one point on $\mathcal{K}$. Such divergence of VP from unimodality can be controlled with the prior.

So consider two normal probability measures $P$ and $Q$ on $\mathbb{R}^2$, with parameters $\theta_p=(\mu_p,\Sigma_p)$ and $\theta_q=(\mu_q,\Sigma_q)$ and denote their PFs onto $\mathcal{K}$ by $f$ with $P^*$ and $Q^*$. Both have PDFs in $\mathcal{K}$, $p^*$ and $q^*$, which, at any point on $\mathcal{K}$, equal sums of PDFs of $P$ and $Q$ on $\mathbb{R}^2$, $p$ and $q$, at all pre-images of that point. 

With that one constructs a particular TopoVAE, that we'll refer to as KleinVAE, as follows:
\begin{itemize}
    \item latent space: $\mathcal{Z}=([0,1]^2/\!\equiv)\cong\mathcal{K}$,
    \item PDF of VP (of latent $z$ given observed $x$): $q^*(z|x)$ -- PF of $\mathcal{N}(\mu_q,\Sigma_q)$ by $f$ (see Eq.~\ref{eq:reparam-through-covering-Klein-covering-map}),
    \item PDF of prior (of latent $z$, unconditioned by $x$): $p^*(z)$ -- PF of $\mathcal{N}(\mu_p,\Sigma_p)$ by $f$ (see Eq.~\ref{eq:reparam-through-covering-Klein-covering-map}).
\end{itemize}

On a forward pass, the encoder, given an input batch of data $x$, outputs the parameters $\theta_q$ of the VP $q^*$. By definition of PF, sampling a RV $z\sim f_*q$ can be done by first sampling from $q$ and then applying $f$ to the result. This random sample $z\sim q^*_{\theta_q}$ is then passed to the decoder that outputs the parameters of $p(x|z)$ -- the likelihood of reconstruction $x$ given latent $z$. In turn, the gradient of evidence lower bound (ELBO): 
\begin{equation}
    \textrm{ELBO}(W)=\mathbb{E}_{z\sim q^*_{\theta_q(W_\textrm{enc})}}\,p_{W_\textrm{dec}}(x|z)\\
    -\textrm{KL}(q^*_{\theta_q(W_\textrm{enc})}||\,p^*_{\theta_p})
    \label{eq:reparam-through-covering-Klein-ELBO}
\end{equation}
w.r.t. $W=(W_\textrm{enc},W_\textrm{dec})$ -- the parameters of the VAE network's layers is easily backward passed. The above mentioned inequalities state that KL-divergence between $q^*$ and $p^*$ on $\mathcal{Z}=\mathcal{K}$ is bounded from above by KL between $q$ and $p$ on $\mathbb{R}^2$ -- so the negative KL term in ELBO (\ref{eq:reparam-through-covering-Klein-ELBO}) is bounded from below by
\begin{equation}
    -\textrm{KL}_\mathcal{K}(q^*_{\theta_q}||\,p^*_{\theta_p})
    \geq -\textrm{KL}_{\mathbb{R}^2}(q_{\theta_q}\,||\,p_{\theta_p}),
    \label{eq:reparam-through-covering-Klein-KL-bound}
\end{equation}
-- KL-div between two bivariate Gaussians $q_{\theta_q}$ and $p_{\theta_p}$, \CORRECTION{for which a well-known analytic formula exists}:
\begin{equation}
\begin{aligned}
\textrm{KL}_{\mathbb{R}^2}(q\,||\,p)=\tfrac{1}{2}\big(\log\tfrac{|\Sigma_p|}{|\Sigma_q|}-2+\textrm{tr}(\Sigma_p^{-1}\Sigma_q)+\\
+(\mu_p-\mu_q)^T\Sigma_p^{-1}(\mu_p-\mu_q)\big)
\end{aligned}
\end{equation}

\CORRECTION{We thus replace the KL between pushforwards with this analytic expression, maximizing the resulting lower bound on (\ref{eq:reparam-through-covering-Klein-ELBO}). Although this bound is potentially loose, this is computationally advantageous, mitigating Monte-Carlo estimation of KL between pushforwards.}

So what's left is to define how to backward-pass the gradient (w.r.t. $\theta_q$) through sampling (in the first, reconstruction loss, term of ELBO):
\begin{equation}
    z\sim q^*_{\theta_q}=f_*\,q_{\theta_q}=f_*\,\textrm{PDF}_{\mathcal{N}(\mu_q,\Sigma_q)}
    \label{eq:reparam-through-covering-Klein-sampling}
\end{equation}
-- this stochastic map is decomposed into sampling a standard RV $\varepsilon\sim\mathcal{N}(0,I)$, and then then reparameterizing it
\begin{equation}
    z=f(\mu_q+\Sigma_q^{1/2}\,\varepsilon).
    \label{eq:reparam-through-covering-Klein-sampling-decomposition}
\end{equation}
This provides gradient passage through reparameterization via covering, since the $f$ map is a covering map, being part of the reparameterization map (\ref{eq:reparam-through-covering-Klein-sampling-decomposition}).
With the covering being (\ref{eq:reparam-through-covering-Klein-covering-map}) simply a sheet-wise identical projection (albeit changing the orientation on half of the sheets), differentiating it is trivial, and can be safely delegated to automatic differentiation -- as $f$ is only discontinuous at the boundaries of the sheets, which are a set of Lebesgue measure zero in $\mathbb{R}^2$. Practically it is thus enough to implement $f$ with e.g. differentiable \texttt{torch.where} and  \texttt{torch.remainder} functions of Pytorch library \cite{pytorch}.

\section{Experiment, KleinVAE}

\subsection{Artificial Klein-Circles dataset}

As a proof of validity of our method, we implement a VAE with the latent space having the topology of the Klein bottle, that we call KleinVAE. We experimented with learning the distribution of Gabor-Klein filters (see Fig. \ref{fig:diagrams-of-filters}) with it, but found it not so easy with shallow autoencoders with standard layers -- possibly due to the non-linear oscillating nature of these filters. That is why we provide a demo of KleinVAE on some other artificial data.

\begin{figure}[h!]
  \centering
  \includegraphics[width=\linewidth]{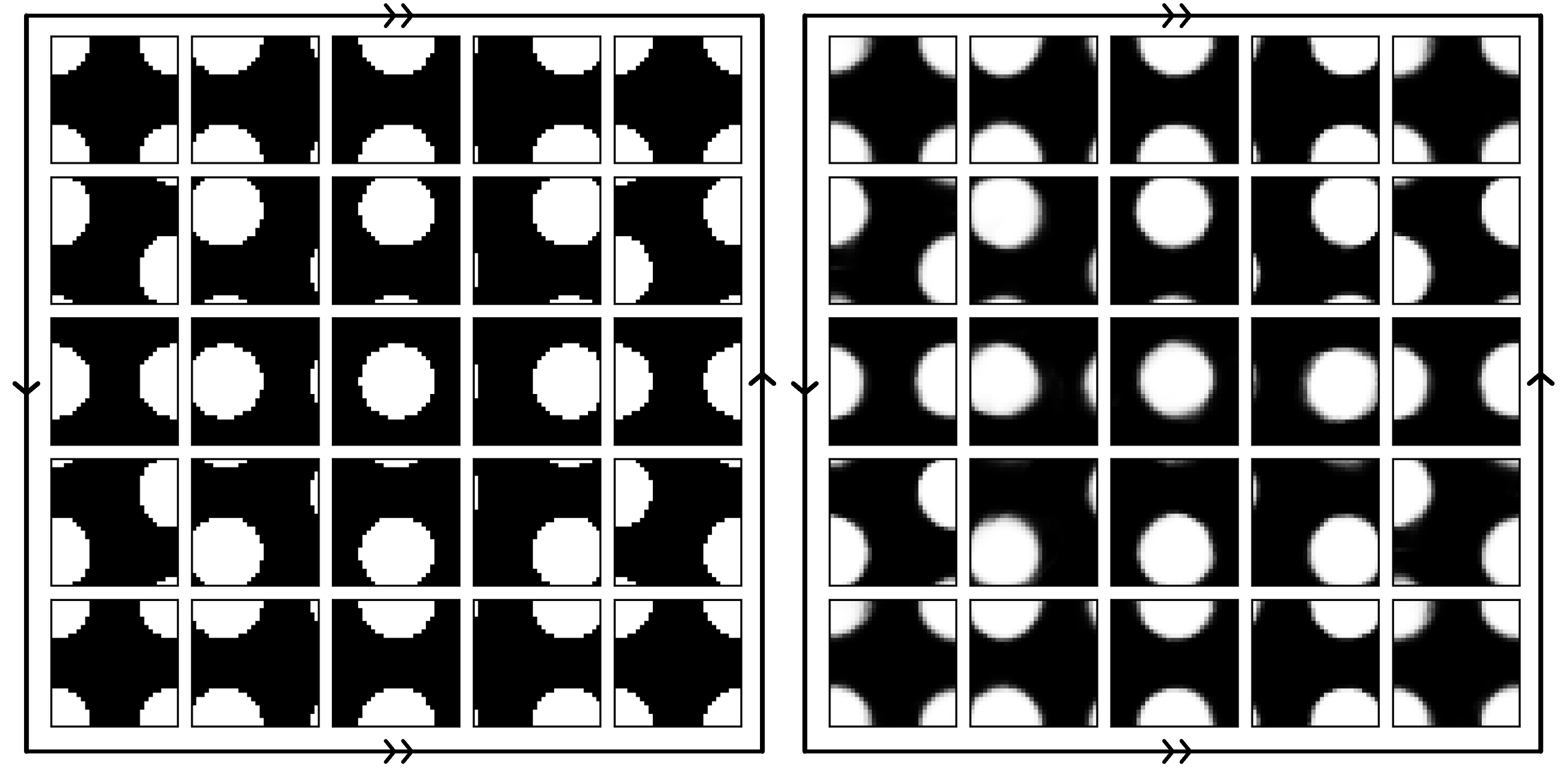}
  \caption{Klein-Circles dataset. \textbf{Left}: samples of original data -- circles on the Klein bottle with centers at different positions. \textbf{Right}: reconstructions of corresponding images with KleinVAE. White = intensity equals 1, black = zero.}
  \label{fig:circles}
\end{figure}

We generate small images ($30\!\times\!30$ pixels) of a circle of radius $0.3$ on a unit square. Boundary points of the square are identified in such a way that it has topology of $\mathcal{K}$ (see Fig. \ref{fig:circles}). With circle center at uniformly random positions, this forms our Klein-Circles dataset. We generate $100.000$ such images.

\begin{figure}[h!]
  \centering
  \includegraphics[width=\linewidth]{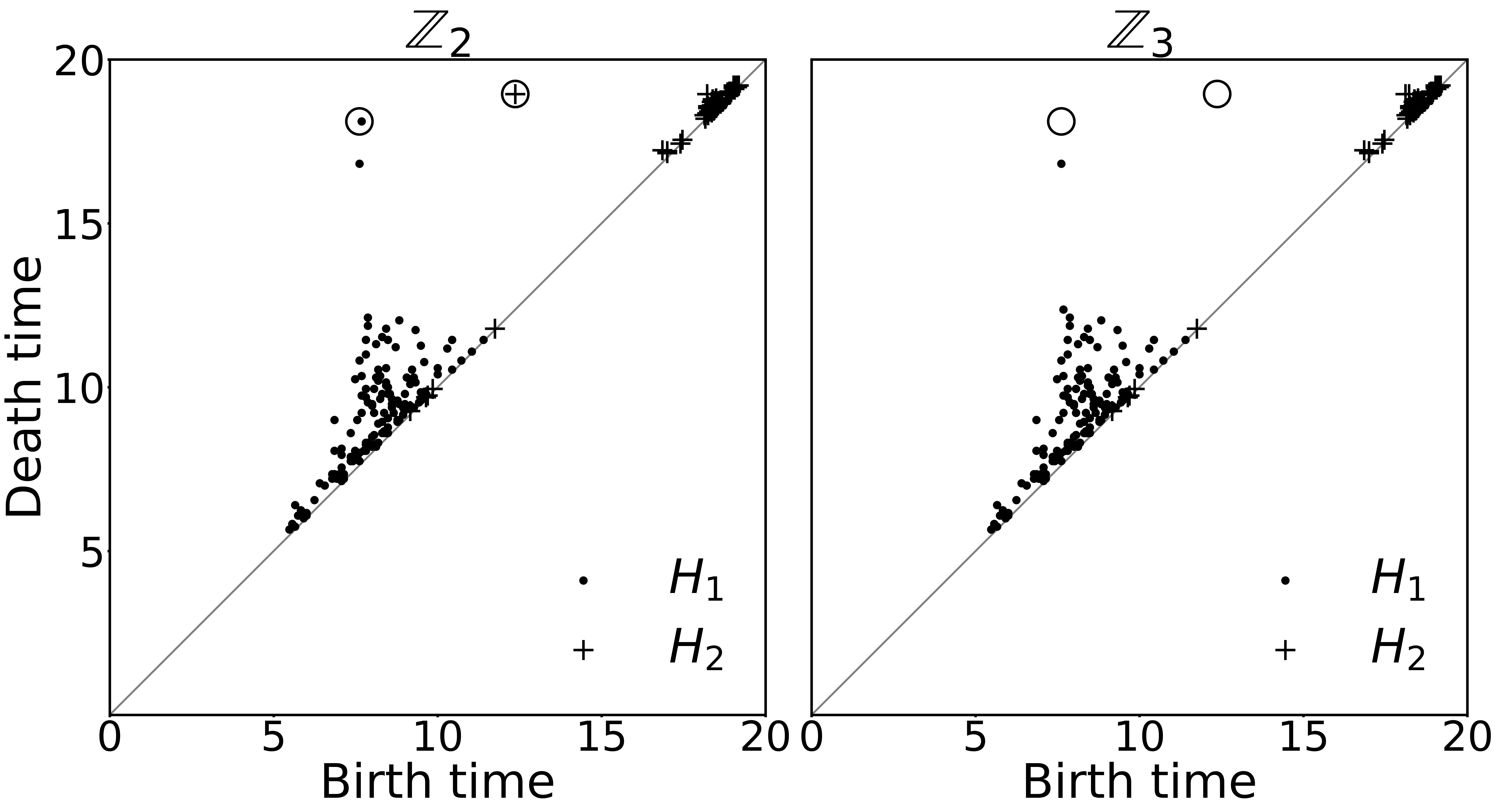}
  \caption{Persistence diagrams of 500 images decoded by KleinVAE from images with uniformly random circle center positions.}
  \label{fig:diagram-of-circles-KleinVAE}
\end{figure}

We train a KleinVAE with the encoder made of fully-connected layers of sizes $30\!\times\!30=900\!\to\!1024\!\to\!512\!\to\!128\!\to\!32\!\to\!5$ (so the latent space has dimension 5, since on it we predict 5 parameters of the variational posterior (VP): 2 components of the $\mu$ vector and $3$ components of the lower-triangular scale matrix, that, squared, provides the $2\!\times\!2$ covariance matrix of the VP), and the decoder with similar ones but in reverse sequence, with Leaky-ReLU nonlinearity. Batch size is set to 1024, training is done with Adam optimizer \cite{kingma2014adam} for 300 epochs with initial learning rate $\textrm{lr}=10^{-3}$ with LR scheduler \texttt{ReduceLROnPlateau} with reduction factor of $0.99$. \CORRECTION{The prior is set to a (pushforward onto $\mathcal{K}$ of) normal with $\mu=(0.5,0.5)$ (center of the square) and $\Sigma^2=\textrm{diag}(0.1,0.1)$. Random seed is set to 27.} 

We compute persistent homology of reconstructions of a sample of $500$ images with uniformly random circle center positions -- persistence diagrams can be seen on Fig. \ref{fig:diagram-of-circles-KleinVAE}. These suggest that KleinVAE successfully learns the latent structure of the Klein bottle.

\begin{figure}[h!]
  \centering
  \includegraphics[width=\linewidth]{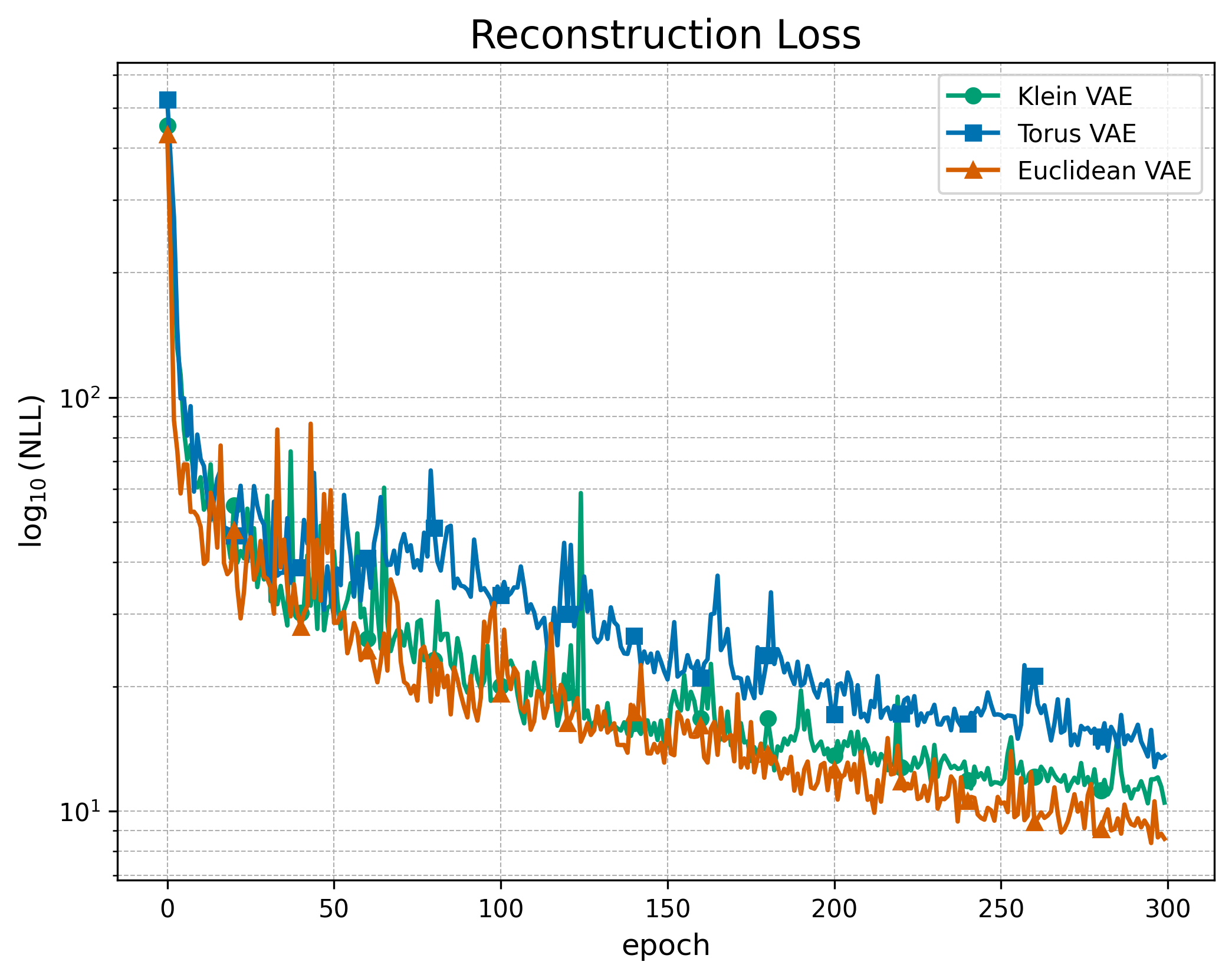}
  \caption{Reconstruction loss (pixel-summed Bernoulli negative log-likelihood per observation) training dynamics.}
  \label{fig:reconstruction-loss}
\end{figure}

\CORRECTION{We compare KleinVAE to a VAE with the latent space being a torus (only $f_{\mathbb{R}^2/\mathbb{Z}^2}$ covering being part of the reparameterization map), as well as to a ``vanilla'' VAE with the latent space being whole 2-dimensional Euclidean space. Other architectural and training details of these models are set to be the same. Figure \ref{fig:reconstruction-loss} shows the training dynamics: KleinVAE reaches better reconstruction than toric VAE, but worse than unrestricted Euclidean VAE.}

\CORRECTION{We also track a topological quality metric for all three models: bottleneck distance between persistence diagrams of a subsample of 400 random images and their reconstructions. Bottleneck distance between diagrams $X$ and $Y$ -- point-sets in the 2D plane -- is $\infty$-Wasserstein distance:}
\begin{equation}
    W_\infty(X,Y)=\inf_{\eta:X\to Y} \sup_{x\in X}||x-\eta(x)||_\infty
\end{equation}
\CORRECTION{-- where the infimum is taken over all matchings (to ensure a perfect matching always exists, one can match points from $X$ and $Y$ to points on the diagonal of the PD).}
\begin{figure}[h!]
  \centering
  \includegraphics[width=0.9\linewidth]{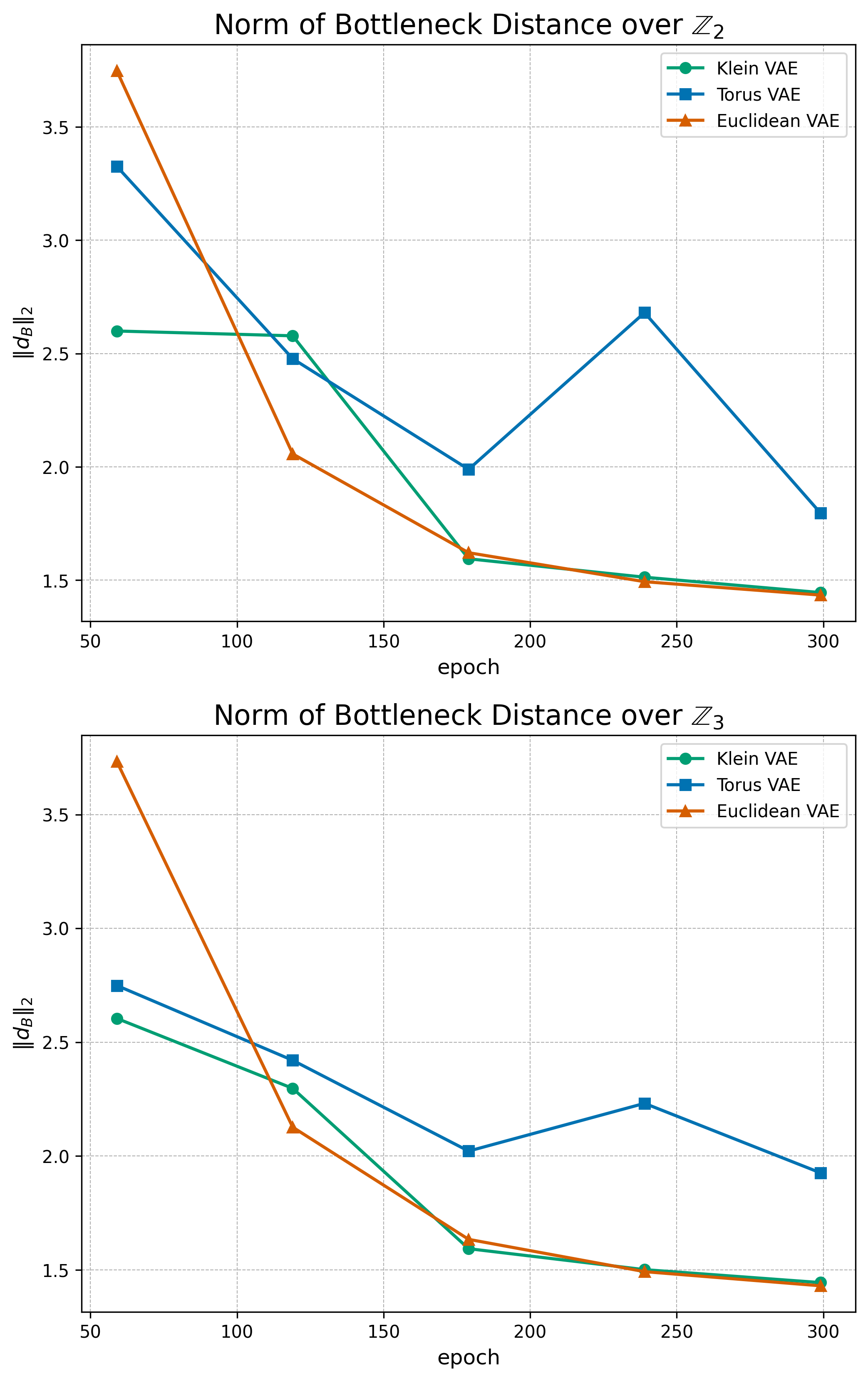}
  \caption{Topological score dynamics.}
  \label{fig:topological-score}
\end{figure}

\begin{figure}[h!]
  \centering
  \includegraphics[width=\linewidth]{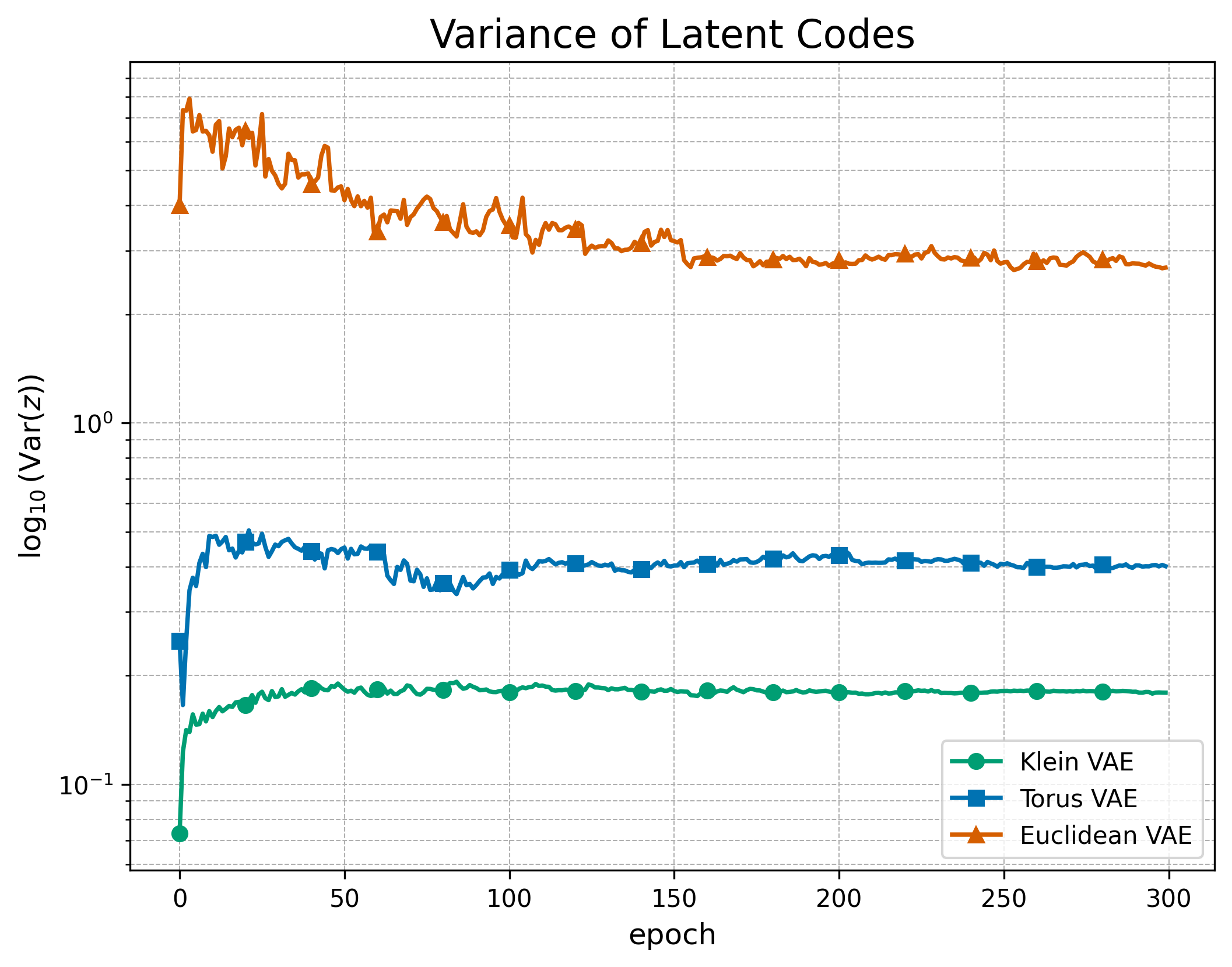}
  \caption{Empirical variance of latent codes dynamics.}
  \label{fig:var-z}
\end{figure}

\CORRECTION{Figure \ref{fig:topological-score} demonstrates that KleinVAE starts with a better topological score than other models, and eventually is not beaten by the Euclidean VAE, with the latter having more freedom to encode data, which is confirmed by significantly higher empirical variance of latent codes (see Fig. \ref{fig:var-z}) -- so in this simple experiment a topology-informed model can be said to do as well with fewer resources.}

\section{Future applications}

With the main claimed contribution of present work being a generalization of RT -- with reparameterization via coverings shown possible, advocated to be used in generative modeling (not limited to VAEs -- seen as basic latent models that can be used as building blocks for more expressive ones to figure out first), we're not merely interested in this technique per se. We hope to highlight connections seen between several directions of research: gen. modeling, Bayesian learning, and topological data analysis (TDA) -- by proposing an exemplar construction of such a topology-aware (manifold-supported) generative models, that can be used as a prior (in the sense of Bayesian inference) on the parameters of some other model, thus providing it with relevant topological inductive bias.

\subsection{Bayesian learning}

Bayesian inference in machine learning, or just Bayesian learning (BL) -- is a motivation of present work. BL is developing since at least 90s \citep{Bayesian-learning-Neal,Bayesian-learning-Neapolitan}, and is still relevant in the modern age of deep learning (DL) \citep{Bayesian-DL,Bayesian-DL-priors}.

Any model processing data $X$ is specified by its parameter $\theta$ -- if the model is a neural network (NN), $\theta$ contains the weights of its layers. In BL, instead of considering one optimal parameter value given training data $X$, one considers
\begin{equation}
    p(\theta|X)=\frac{p(X|\theta)\,p(\theta)}{p(X)}
    \label{eq:intro-Bayes-Bayes-formula}
\end{equation}
the whole posterior probability distribution $p(\theta|X)$ of possible parameter $\theta$ values given observed data $X$. From Bayes' theorem (\ref{eq:intro-Bayes-Bayes-formula}), it follows that $p(\theta|X)$ is proportional to the likelihood $p(X|\theta)$ of observed data $X$ given a certain value of $\theta$ -- times the prior $p(\theta)$ probability of parameter $\theta$ taking a certain value, independent of observed data. With a prior $p(\theta)$ one can incorporate all of knowledge of what values $\theta$ should or should not likely take -- into the model. Typically, such Bayesian models are trained with Monte-Carlo (MC) techniques, for which it is enough to be able to sample random values $\theta\sim p(\theta)$ from the prior. 

In principle, incorporating informative prior beliefs into models improves their performance and generalization ability (if the prior is ``strong'' enough compared to the amount of training data available). We were particularly motivated by weight priors for vision models: it was shown in \cite{Priors-for-CNNs} that imposing priors on the structure of the filters  of convolutional layers of CNNs (the original LeNet \citep{CNN-beginnings-LeNet}, VGG \citep{CNN-VGG}, ResNet \citep{CNN-ResNet})) improves their performance (faster training to reach same accuracy) and generalization ability (higher entropy of predictions of an untrained CNN initialized only from the prior); see also \citep{fortuin2021bayesian}. Said imposed structure is, in fact, well-known -- filters are sampled from the family of Gabor filters \citep{Gabor-filters} (rather than from uncorrelated Gaussian noise components), that were widely used in CV before the success of CNNs. This ``revival'' of Gabor filter usage -- as weight initialization for CNNs -- happened before: \citep{pre-GaborNet,GaborNet}, but it was in \cite{Priors-for-CNNs} where this was framed as Bayesian learning.

Further on, it was proposed in \cite{DWP} to learn (with a generative model, e.g. a VAE) such distributions of useful NN weights (CNN filters) from a NN trained on one dataset -- to transfer those as priors for NNs operating on other datasets -- this was called Deep Weight Prior (DWP), following the introduction of Deep Image Prior (DIP) in \cite{DIP}.

Said works strongly motivated present work, which advocates paying more attention to topological structures when doing probabilistic modeling, as, in particular, in \cite{jin2024implications}. As described further, one such a topological structure for CNN weights in particular has been known for some time, so we propose to combine the flexibility of priors learned with a generative model (e.g. a VAE) with such ``topological inductive bias'': knowing which exact manifold the prior distribution is in fact supported on. We propose to refer to such priors as Topological Weight Priors.

\subsection{Priors for Topological Deep Learning: Klein bottle in vision}

Another side of this, seemingly, single direction of research -- is topological deep learning (TDL). Its motivating premise is that data possesses certain topological structure (e.g due to restrictions of symmetry) to account for when designing models processing such data \citep{Topology-and-data}.

One particular observation that inspired much development in the field of topological data analysis (TDA) -- is that such structure is found in natural images \citep{Nonlinear-statistics-patches}: if one considers grayscale images of natural objects at a small scale -- i.e. (high contrast) patches of e.g. $3\!\times\!3$ pixels, the distribution of such patches would not fill the entire feature space $\mathbb{R}^9$, but would rather concentrate near a certain (embedded) manifold of dimension 2. Using the construction of persistent homology \citep{Barannikov,Robins,Edelsbrunner-Zomorodian,Zomorodian-Carlsson}, the vehicle of TDA, it was shown in \cite{Klein-bottle-natural-images} that said manifold, in fact, has a topological structure very similar to that of the Klein bottle~$\mathcal{K}$.

It was proposed in \cite{Klein-bottle-texture-dict} to utilize this knowledge to build low-parameter dictionaries of natural textures. More recently, it was proposed in \cite{Gabrielsson,Gabrielsson-2,Topo-Conv-Layers} to directly incorporate said topological structure into CNNs -- authors refer to these as Topological CNNs (TCNNs). The idea is clear: by design \citep{CNN-beginnings-LeNet}, convolutional filters of CNNs are meant to detect repeating motifs of small image patches, activating most on patches they are most co-aligned with -- so if certain structure is present in typical image patches, a similar, dual structure should emerge in the filters of a trained CNN, this is exactly what \cite{Gabrielsson} observed. These filters are, in fact, a certain subset of the mentioned Gabor filters -- to which we refer to as Gabor-Klein filters.

The proposal of \cite{Topo-Conv-Layers} with TCNNs is thus to incorporate these Gabor-Klein filters into (at least, early) layers of CNNs by sampling these from their (Klein bottle) topology. That is done essentially by discretizing this topology into a graph and sampling from a (discrete) distribution on it. In contrast, our proposal is to learn a smooth distribution of informative filters on a smooth representation of such a priori known topologies -- with an appropriate generative model.

\section{Conclusion}

In present work, we show how one can do the reparameterization trick of variational inference on various topologically non-trivial spaces. This allows to use these as latent spaces of topology-aware generative models, as we exemplify by constructing a Topo-VAE with the latent space having the topological structure of the Klein bottle. This is novel since, to date, reparameterization for such non-trivial topologies has been explicitly generalized, of comparably non-trivial cases -- only to Lie groups. \CORRECTION{We propose to construct reparameterization maps via covering maps, which allow for arbitrarily complicated sets of discontinuities -- provided these have measure zero. We establish an inequality on KL-divergence between the variational posteriors on the base of the covering, bounding it by KL-divergence between their pullbacks on the cover, which is useful if the latter has an analytic expression, or is elsehow computationally advantageous for ELBO estimation.} We construct such a TopoVAE with latent Klein bottle topology and report its performance on an appropriate artificial dataset. We also speculate on applicability of such topology-aware generative models as model weight priors in Bayesian learning, providing potentially helpful topological inductive bias.

\section*{Code and data availability}
Supplementary code is available at \url{https://github.com/bekemax/KleinVAE}.

\section*{Acknowledgements}
The work of M. Beketov is an output of a research project HSE-BR-2025-001 implemented as part of the Basic Research Program at HSE University. We thank Anton Ayzenberg, Ilya Schurov, Ivan Gonchar and Vassily Gorbounov for fruitful discussions.

\bibliographystyle{unsrt}
\bibliography{OUR-BIB}

\end{document}